\documentclass[12pt]{article}
 \usepackage{multirow}
\usepackage{fullpage}

\usepackage{algorithm, algorithmic}
\usepackage{amsmath, amssymb}
\usepackage{graphicx}
\usepackage{verbatim}
\usepackage{color}
\definecolor{DarkGreen}{rgb}{0.1,0.5,0.1}
\definecolor{DarkRed}{rgb}{0.5,0.1,0.1}
\definecolor{DarkBlue}{rgb}{0.1,0.1,0.5}
\usepackage{enumitem}
\newcommand \bD{\mathbf{D}}
\newcommand \bS{\mathbf{S}}
\newcommand \bU{\mathbf{U}}
\newcommand \bV{\mathbf{V}}
\newcommand \bM{\mathbf{M}}
\newcommand \bW{\mathbf{W}}
\newcommand \bR{\mathbf{R}}

\newcommand \bX{\mathbf{X}}
\newcommand \bL{\mathbf{L}}

\newcommand \bu {\mathbf{u}}

\newcommand \bg {\mathbf{g}}
\newcommand \bw {\mathbf{w}}
\newcommand \bx {\mathbf{x}}

\newcommand{\cF}{\mathcal{F}}

\newcommand{\cI}{\mathcal{I}}

\newcommand{\cL}{\mathcal{L}}

\newcommand{\cU}{\mathcal{U}}

\newcommand{\cX}{\mathcal{X}}

\newtheorem{theorem}{Theorem}[section]

\newtheorem{remark}[theorem]{Remark}

\newcommand{\dd}[2] { \langle {#1}, {#2} \rangle}

\newcommand{\mtx}[1] { \mathbf {#1}}

\newcommand {\diag} {\rm{diag}}
\newcommand {\alg} {\texttt{S$^2$COR}}

\newcommand{\dsML}{ML-IMDB}
\newcommand{\dsAM}{Amazon}
\newcommand{\dsCU}{CiteULike}

\floatstyle{boxed}

\usepackage{mathrsfs}
\begin{document}

\title{Semi-supervised  Collaborative Ranking with\\ Push at  Top}

\author{Iman Barjasteh$^{\dagger *}$,~Rana Forsati$^{\ddagger}$\footnote{These authors contributed equally to this work.},~Abdol-Hossein~Esfahanian$^{\ddagger}$,~Hayder~Radha$^{\dagger}$\\ \\
{\small$^{\dagger}$Department of Electrical and Computer Engineering} \\
\small{$^{\ddagger}$Department of Computer Science and Engineering}\\
\small{Michigan State University}\\\\
\small{\texttt{\{forsati,esfahanian\}@cse.msu.edu,~\{barjaste,radha\}@msu.edu}}}


\date{}

\maketitle


\begin{abstract} 

Existing collaborative ranking based recommender systems tend to perform best when there is enough observed ratings for each user and the observation is made completely at random. Under this setting recommender systems can properly suggest a list of recommendations according to the user interests. However, when the observed ratings are extremely sparse (e.g. in the case of cold-start users where no rating data is available), and are not sampled uniformly at random, existing ranking methods fail to effectively leverage side information to transduct the knowledge from existing ratings to unobserved ones. We propose a \textbf{semi-supervised collaborative ranking} model, dubbed \alg, to  improve the quality of cold-start recommendation. \alg~mitigates the sparsity issue by leveraging side information about \textit{both observed and missing} ratings by collaboratively learning the ranking model. This enables it to deal with the case of missing data not at random, but to also effectively incorporate the available side information in transduction. We experimentally evaluated our proposed algorithm on a number of challenging real-world datasets and compared against state-of-the-art models for cold-start recommendation. We report significantly higher quality recommendations with our algorithm compared to the state-of-the-art.
\end{abstract}



\section{Introduction}

Due to the popularity and exponential growth of e-commerce  and online streaming websites, a compelling demand has been created for efficient recommender systems to guide users toward items of their interests (e.g. products, books, movies)~\cite{100}. In collaborative methods, either \textit{filtering} or \textit{ranking}, by relying on the low-rank assumption on the users' preferences, both users and items are mapped into a latent feature space based on partially observed ratings that are later used to make predictions. In collaborative filtering (CF) methods such as matrix factorization~\cite{koren2009matrix}, where the aim is to accurately predict the ratings, the latent features are extracted in a way to minimize the prediction error measured in terms of popular performance measures such as root mean square error (RMSE). In spark contrast to CF,   in collaborating ranking (CR) models~\cite{koren2009matrix,christakopoulou2015collaborative,volkovs2012collaborative,cremonesi2010performance}, where the goal is to rank the unrated items in the order of relevance to the user,  the popular ranking measures such as  as discounted cumulative gain (DCG), normalized
discounted cumulative gain (NDCG), and average precision (AP)~\cite{jarvelin2000ir}  are  often employed to collaboratively learn a ranking model for the latent features.

Recent studies have demonstrated  that CR models lead to significantly higher ranking
accuracy  over their traditional CF counterparts  that optimize rating prediction. This  is important considering the fact that what we really care in recommendation  is not the actual values of ratings, but the order of items to be recommended to a specific user.  Therefore, the error  measures such as RMSE are often hopelessly insufficient,  as their place equal emphasis on all the ratings.  Among ranking models,  the methods that mainly concentrate on the \textit{top of the list} have received a considerable amount of attention, due to the higher probability of examining the top portion of the list of recommendations by users.  Therefore, the introduction of ranking metrics such as push norm or infinite norm~\cite{rudin2009p,agarwal2011infinite,christakopoulou2015collaborative,lee2014local},  sparked a widespread interest in CR models and has been proven to be more effective in 
practice~\cite{steck2015gaussian,christakopoulou2015collaborative}.

%

Although CR models for recommender systems has  been studied extensively and some progress has been made, however,  the state of affairs remains unsettled:  the issue of handling \textit{cold-start} items in ranking models  and coping with \textit{not missing at random} assumption of ratings  are  elusive open issues. First, in many real world applications, the rating data are very sparse (e.g., the density of the data is around 1\% for many publicly available datasets) or for a subset of users or items the rating data is entirely missing (knows as cold-start user and cold-start item problem, respectively)~\cite{schein2002methods}. Second, collaborative filtering and ranking models rely on the critical assumption that  the  missing ratings are sampled uniformly at random. However, in many real applications of recommender systems, this assumption is not believed to hold,  as invariably some users are more active than others and some items are rated by many people while others are rarely rated~\cite{steck2010training}.  These issues have been  investigated in factorization based methods,  nonetheless, it is not straightforward to adapt them to CR models and are  left open~\cite{christakopoulou2015collaborative}. Motivated by these challenges, we ask the following fundamental question in the context of collaborative ranking models:
\begin{quote}
\textit{Is it possible to effectively learn a collaborative ranking model in the presence of cold-start items/users that is  robust to  the sampling of observed ratings?}
\end{quote}

In this paper, we give an affirmative answer to the above question.  In particular, we introduce a   \textit{semi-supervised collaborative ranking} model, dubbed~\alg~,  by leveraging side information about \textit{both observed and missing ratings} in collaboratively learning the ranking model.  In the learned model, unrated items are conservatively pushed after the relevant  and before the irrelevant items in the ranked list of items for each individual user. This crucial difference greatly boosts the performance and limits the bias caused by learning only from sparse non-random observed ratings. We also introduce a graph regularization method to exploit the side information about users to overcome the cold-start users problem.  In summary, the key features of~\alg~are:

\begin{itemize}[noitemsep,topsep=10pt,parsep=5pt,partopsep=5pt]
\item Inspired by recent developments in ranking at top~\cite{rudin2009p,agarwal2011infinite,li2014top}, the proposed model is a collaborative ranking model   that   primarily focuses on  the top of the recommendation  list for each user. Moreover, in stark contrast to pairwise ranking models which have quadratic dependency on the number of items, the proposed ranking model has a linear dependency on the number of items, making it suitable for large-scale recommendation.
\item It leverages side information about items with both observed and missing ratings while collaboratively learning the ranking model, which enables it  to effectively incorporate the available side information in knowledge transduction.
\item By incorporating the unrated items in ranking, it limits the bias caused by learning solely based on the observed ratings and consequently deals with the  not missing at random issue of ratings.
\item It is also able to leverage the similarity information between users based on a graph regularization method to make high quality recommendations for users with few ratings or cold-start users without an rating information.
\end{itemize}

To build the intuition on how incorporating missing ratings in~\alg~is beneficial in handling cold-start problem and mitigating data sparsity issue,  we note that  in many real world applications the available feedback on items is extremely sparse, and therefore the ranking models fail to effectively leverage the available side information in transdcuting the knowledge from existing ratings to unobserved ones. This problem becomes especially eminent in cases where surrogate ranking models such as pairwise models are used due to their computational virtues, where the unobserved ratings do not play any role in learning the  model.  As a result,  by leveraging rich sources of information about  all items, one can potentially bridge the gap between existing items  and new  items  to overcome the cold-start problem.

Turning to the non-random sampling issue of observed ratings, we note that the non-randomness is observing the ratings creates a  bias in  learning the model that negatively impacts the future predictions and may degrade the resulting recommendation accuracy if ignored.  Therefore, the nature of missing ratings  has to be modeled precisely as to obtain correct results. To reduce the effect of bias, the proposed ranking model takes a conservative approach and pushes the items with unknown ratings to the middle of ranked list, i.e., after the relevant and before the irrelevant items. This is equivalent to assuming a prior about the unknown ratings which is believed  to perform well  as investigated in~\cite{devooght2015dynamic}. However, unlike~\cite{devooght2015dynamic}, the proposed ranking idea is free of deciding an explicit value for  missing ratings which makes it more valuable from a practical point of view.

We conduct thorough experiments on real datasets and compare our results with the state-of-the-art models for cold-start recommendation to demonstrate the effectiveness of our proposed algorithm in recommendation at the top of the list and mitigating the data sparsity issue. Our results indicate that our algorithm outperforms other algorithms and provides recommendation with higher quality compared to the state-of-the-art methods. \\

\noindent\textbf{Organization.} {This paper is organized as follows. We briefly
 review related work in Section~\ref{sec:related}. We establish the notation and formally define the problem in Section~\ref{sec:preliminaries}. In Section~\ref{sec:taco}, we propose the semi-supervised collaborative ranking model with a push at the top of the list. Section~\ref{sec:optimization} discusses  efficient convex and non-convex optimization algorithms for optimization. Section~\ref{sec:taco-users} generalizes the proposed algorithm to leverage similarity information about users. We empirically evaluate the proposed method in Section~\ref{sec:results}, and conclude in Section~\ref{sec:conclusions}.}

\section{Related Work}\label{sec:related}

There is now a vast body of literature on ranking models for recommendation, coping with non-random missing ratings and an even bigger body of literature on handling cold-start problem by leveraging side information; we restrict our literature review here to papers that are most directly related.

\noindent\paragraph{Collaborative ranking for recommendation.~}{The last few years have seen a resurgence in collaborative ranking   centered around the technique of exploiting low-rank structures, an approach we take as well. Several approaches to CR have recently been proposed that are mainly inspired by the analogy between query-document relations in IR and user-item relations in recommender systems. The PMF-based approach~\cite{balakrishnan2012collaborative} uses the latent representations produced by matrix factorization as user-item features and learns a ranking model on these features. CofiRank~\cite{weimer2007maximum} learns latent representations that minimize a ranking-based loss instead of the squared error. ListRankMF~\cite{shi2010list}  aims at minimizing the cross entropy between the predict item permutation probability and true item permutation probability. In~\cite{lee2014local} a method for Local Collaborative Ranking (LCR) where ideas of local low-rank matrix approximation were applied to
the pairwise ranking loss minimization framework is introduced. In~\cite{steck2015gaussian} a  framework that allows for pointwise as well as listwise training with respect to various ranking
metrics is proposed. Finally,~\cite{christakopoulou2015collaborative} proposed a CR model build on the recent developments in ranking methods~\cite{agarwal2011infinite,rudin2009p} that focus on accuracy at top and proposed CR methods with p-push and infinite push norms. Incorporating side information in their model which is left as open issue was the main motivation of the current work.
}

\paragraph{Cold-start recommendation with side information.~}{Due in part to its importance,  there has been an active line of work to address difficulties associated with cold-start users and items, where a  common  theme  among  them  is  to  exploit  auxiliary  information  about  users  or  items  besides  the  rating data  that  are  usually  available~\cite{shi2014collaborative}. A feature based regression ranking model for predicting the values (rates) of user-item matrix in cold-start scenarios by leveraging all information available for users and items is proposed in~\cite{park2009pairwise}. The kernelized matrix
factorization approach studied in~\cite{zhou2012kernelized}, which incorporates the auxiliary information into the MF. In~\cite{saveski2014item} joint factorization of the user-item and item-feature matrices by using the same item latent feature matrix in both decompositions is utilized. The FBSM model is introduced in~\cite{sharmafeature2015b}, which learns factorized
bilinear similarity model for new items, given the rating information as well as the features of these items. Recently,~\cite{barjasteh2015cold} proposed a decoupling approach to transduct knowledge from side information to rating prediction which is able to handle both cold-start items and users problems in factorization based models.

}

\noindent\paragraph{Recommendation with not missing at random ratings.~}{
Substantial evidence for violations of the missing at random condition in recommender systems is reported in~\cite{missing-at-random2007} and it has been  showed that incorporating an explicit model of the missing data mechanism can lead to significant improvements in prediction performance.The first study of the effect of non-random missing data on collaborative ranking is presented in~\cite{marlin2009collaborative}. In~\cite{sindhwani2010one}  an EM algorithm to optimize in turn the factorization and the estimation of missing values. Recently, in~\cite{devooght2015dynamic} a novel dynamic matrix factorization framework that allows to set an explicit prior on unknown values is introduced. However their algorithm requires a careful setting of the prior rating to be practical.}




\section{Preliminaries}\label{sec:preliminaries}

In this section we establish the notation used throughout the paper and formally describe our problem setting. 

Scalars are denoted by  lower case letters and vectors by  bold face lower case letters such as $\bu$. We use  bold face upper case letters such as $\bM$ to denote matrices.   The Frobenius norm of a matrix $\bM \in \mathbb{R}^{n \times m}$ is denoted by $\|\bM\|_\text{F}$,  i.e,     $\|\mathbf{M}\|_{\rm{F}}=\sqrt{\sum_{i=1}^n\sum_{j=1}^m |M_{ij}|^2}$ and its $(i,j)$th entry is denoted by $A_{i,j}$. The trace norm of a matrix is denoted by $\|\bM\|_{*}$ which is defined as the sum of its singular values. The transpose of  a vector and a matrix denoted by $\bu^{\top}$ and $\bU^{\top}$, respectively. We use $[n]$ to denote the set  on integers $\{1,2, \cdots, n\}$. The set of non-negative real numbers is denoted by $\mathbb{R}_{+}$. The indicator function is denoted by $\mathbb{I}[\cdot]$. For a vector $\bu \in \mathbb{R}^p$ we use $\|\bu\|_1 = \sum_{i=1}^{p}{|\bu_i|}$, $\|\bu\|_2 = \left(\sum_{i=1}^{p}{|\bu_i|^2}\right)^{1/2}$, and $\|\bu\|_{\infty} = \max_{1 \leq i \leq p } \bu_i$ to denote its $\ell_1$, $\ell_2$, and $\ell_{\infty}$ norms, respectively.  The dot product between two vectors $\bu$ and $\bu'$ is denoted by either $\dd{\bu}{\bu'}$ or $\bu^{\top}\bu'$.

In collaborative filtering we assume that there is  a set of $n$ users $\cU = \{u_1, \cdots, u_n\}$ and a set of $m$ items $\cI = \{i_1, \cdots, i_m\}$ where  each user $u_i$ expresses opinions about a set of items. The rating information is summarized in an $n \times m$ matrix $\bR \in \{-1, +1, ?\}^{n \times m}, 1 \leq i \leq n, 1 \leq j \leq m$ where the rows correspond to the users and the columns correspond to the items and $(p,q)$th entry is the  rate given by user $u_p$ to the item $i_q$. We note that the rating matrix is partially observed and it is sparse in most cases.  We are mainly interested in recommending a set of items for an active user such that  the user has not rated these items before.

\section{Transductive Collaborating Ranking}\label{sec:taco}

We now turn our attention to the main thrust of the paper where  we  present  our  transductive collaborative ranking  algorithm  with accuracy at top by exploiting the features of unrated data.  We begin with the basic formulation and then extend it to incorporate the unrated items. The pseudo-code of the resulting learning algorithm is provided in Algorithm~\ref{alg:taco}.

\subsection{A basic formulation}
$\quad$  We consider a ranking problem, where, given a set of users $\cU$ and known user feedback on a set of  items $\cI$, the goal is to generate rankings of  unobserved items, adapted to each of the users' preferences. Here we consider the bipartite setting in which items are either relevant (positive) or irrelevant (negative). Many ranking methods have been developed for bipartite ranking, and most of them are essentially based on pairwise ranking. These algorithms reduce the ranking problem into a binary classification problem by treating each relevant/irrelevant instance pair as a single object to be classified~\cite{liu2009learning}.

As mentioned above, most research has concentrated on the rating prediction problem in CF where the aim is to accurately predict the ratings for the unrated items for each user. However, most applications that use CF typically aim to recommend only a small ranked set of items to each user. Thus rather than concentrating on rating prediction we instead approach this problem from the ranking viewpoint where the goal is to rank the unrated items in the order of relevance to the user. Moreover,  it is desirable  to concentrate aggressively on top portion of the ranked list to include mostly relevant items and  push irrelevant items down from the top. Specifically,   we propose an algorithm that maximizes  the number of relevant items which are pushed to  the absolute top of the list by utilizing the P-Norm Push ranking measure which is specially designed for this purpose~\cite{rudin2009p} . 
 
\begin{algorithm}[t]
\center \caption{\texttt{\alg}}
\begin{algorithmic}[1] \label{alg:taco}
     \STATE \textbf{input:}  
 $\lambda \in \mathbb{R}_{+}$: the regularization parameter, and $\{\eta_t\}_{t \geq 1}$: the sequence of scalar step sizes
	\STATE Initialize $\bW_0 \in \mathbb{R}^{n \times d}$
	\STATE Choose an appropriate step size
	\FOR{$t = 1, \ldots, T$}	
		\STATE Compute the sub-gradient of $\mtx{G}_t \in \partial \cL(\bW_{t})$ using Eq.~(\ref{eqn-grad-u})
		\STATE $[\mtx{U}_t, \mtx{\Sigma}_t, \mtx{V}_t] \leftarrow \texttt{SVD}(\bW_{t-1} - \frac{1}{\eta_{t-1}}\mtx{G}_t))$
		\STATE $\bW_t \leftarrow \mtx{U}_t \left[ \mtx{\Sigma} - \frac{\lambda}{\eta_{t-1}} \mtx{I}\right]_{+} \mtx{V}_t^{\top}$
	 \ENDFOR
    \STATE  \textbf{output:}  
\end{algorithmic}
\end{algorithm}

For simplicity of exposition, let us first consider the ranking model for a single user $u$. Let $\cX^{+} =\{\bx_1^{+}, \cdots, \bx_{n_{+}}^{+}\}$ and $\cX^{-} =\{\bx_1^{-}, \cdots, \bx_{n_{-}}^{-}\}$ be the set of feature vectors of $n_{+}$ relevant and $n_{-}$ irrelevant items to user $u$, respectively.  
We consider linear ranking functions where each  item features  vector $\bx \in \mathbb{R}^d$ is mapped to a score $\bw^{\top}\bx$ . The goal is to find parameters $\bw$ for each user such that the ranking function best captures past feedback from the user. The goal of ranking is to maximize the number of relevant items ranked above the highest-ranking irrelevant item.  We cast this idea for each user $u$ individually into the following optimization problem:
\begin{equation}\label{}
\begin{aligned}
\min_{\bw \in \mathbb{R}^d}  \frac{1}{n^{+}} \sum_{i = 1}^{n^{+}}{\mathbb{I} \left[\dd{\bw}{\bx_i^{+}} \leq \max \limits_{1 \leq j \leq n^{-}} \dd{\bw}{\bx_j^{-}}\right]}
\end{aligned}
\end{equation}
where $\mathbb{I}[\cdot]$ is the indicator function which returns 1 when the input is true and 0 otherwise, $n^{+}$ and $n^{-}$ are the the number of relevant and irrelevant items to user $u$, respectively.

Let us now derive the general form of our objective. We hypothesize that most users base their decisions about items based on a  number of latent features about the items.  In order to uncover these latent feature dimensions, we impose a low-rank constraint on the set of parameters for all users. To this end,  let $\bW = [\bw_1, \bw_2, \cdots, \bw_n]^{\top} \in \mathbb{R}^{n \times d}$ denote the matrix of all parameter vectors for $n$ users. Let $\cI^{+}_{i} \subseteq \{1,2,\ldots, m\}$ and $\cI^{-}_{i} \subseteq \{1,2,\ldots, m\}$ be the set of  relevant and irrelevant items of $i$th user, respectively.     The overall objective for all users is formulated as follows:
\begin{equation}\label{eqn-non-convex-1}
\begin{aligned}
\cF(\bW) &=  \lambda \|\bW\|_{*} \\
&+    \sum_{i = 1}^{n}{\left( \frac{1}{|\cI^{+}_{i}|} \sum_{j \in \cI^{+}_{i}}^{}{{\mathbb{I} \left[\dd{\bw_i}{\bx_j} \leq \max \limits_{k \in \cI^{-}_{i}} \dd{\bw_i}{\bx_k} \right]}} \right)}, 
\end{aligned}
\end{equation}
where $\|\cdot\|_*$ is the trace norm (also known as nuclear norm) which is the sum of the singular values of the input matrix.

The objective in~Eq.~(\ref{eqn-non-convex-1}) is composed of two terms. The first term is the regularization term and is introduced  to capture the factor model intuition discussed above. The premise behind a factor model is that there is only a small number of factors influencing the preferences, and that a user's preference vector is determined by how each factor applies to that user.  Therefore, the parameter vectors of all users must lie in a low-dimensional subspace. Trace-norm  regularization  is  a  widely-used  and  successful  approach  for  collaborative filtering and matrix completion. The trace-norm regularization is well-known to be a convex surrogate to the matrix rank, and has repeatedly shown good performance in practice~\cite{srebro2004maximum,candes2010power}.  The second term is introduced to push the relevant items of each user to the top of the list when ranked based on the parameter vector of the user and features of items.

The above optimization problem is intractable due to the non-convex indicator function. To design practical learning algorithms, we replace the indicator function in~(\ref{eqn-non-convex-1})  with its convex surrogate. To this end, define the convex loss function $\ell: \mathbb{R} \mapsto \mathbb{R}_{+}$ as $\ell(x) = [1-x]_{+}$. This is the widely used hinge loss in SVM classification (see e.g.,~\cite{burges1998tutorial})~\footnote{We note that other convex loss functions such as exponential loss $\ell(x) = \exp(-x)$, and logistic loss $\ell(x) = \log(1+\exp(-x))$ also can be used as the surrogates of indicator function, but for the simplicity of derivation we only consider the hinge loss here.}. This loss function reflects the amount by which the constraints are not satisfied. By replacing the non-convex indicator function with this convex surrogate leads to the following tractable convex optimization problem:
\begin{equation}\label{}
\begin{aligned}
\cF(\bW) &=  \lambda \|\bW\|_{*}  \\
& +    \sum_{i = 1}^{n}{\left( \frac{1}{|\cI^{+}_{i}|} \sum_{j \in \cI^{+}_{i}}^{}{\ell\left(\dd{\bw_i}{\bx_{j}} - \|\bX_i^{-}\bw_i\|_{\infty}\right)} \right)}
\end{aligned}
\end{equation}
where $\bX_i^{-} =[\bx_{1}, \ldots, \bx_{n^{-}_{i}}]^{\top}$ is the matrix of features of $n^{-}_{i}$ irrelevant items in  $\cI^{-}_{i}$ and $\|\cdot\|_{\infty}$ is the max norm of a vector.

\subsection{Semi-supervised collaborative ranking}$\quad$

%

In this part, we extend the proposed ranking idea  to  learn both from rated as well as
unrated items. The motivation of incorporating unrated items comes from the following key observations.  First,  we note that commonly there is a small set of rated (either relevant or irrelevant) items for each user and a large number of unrated items.   As it can be seen from Eq.~(\ref{eqn-non-convex-1}), the unrated items do not play any role in learning the model for each user as the learning is only based on the pair of rated items. When the feature information for items is available, it would be very helpful if one can leverage such unrated items in the learning-to-rank process to effectively leverage the available side information. By leveraging both types of rated and unrated items, we can compensate  for the lack of rating data.  Second, the non-randomness in observing the observed ratings creates a  bias in  learning the model that  may degrade the resulting recommendation accuracy.  Therefore, finding a  precise model  to reduce the effect of bias  introduced by non-random missing ratings seems essential.

To address  these two issues, we extend the basic formulation in Eq.~(\ref{eqn-non-convex-1}) to incorporate items with missing ratings in ranking of items for individual users.  A conservative solution is to  push the items with unknown ratings to the middle of ranked list, i.e., after the relevant and before the irrelevant items. To do so, let $\cI^{\circ}_i = \cI \setminus \left( \cI^{+}_i \cup \cI^{-}_i\right)$ denote the set of items unrated for user $i \in \cU$. We introduce two extra terms in the objective in Eq.~(\ref{eqn-non-convex-1})  to push the unrated items $\cI_{\circ}^{i}$ below the relevant items and above the irrelevant items, which yilelds the following objective:  
\begin{equation}\label{eqn-pnn-indicator}
\begin{aligned}
\cL(\bw)  &=& \frac{1}{|\cI^{+}_{i}|} \sum_{i \in \cI^{+}_{i}}^{}{\ell \left(\dd{\bw}{\bx_i} \leq \max \limits_{j \in \cI^{-}_i} \dd{\bw}{\bx_j}\right)} \\
&+ &  \frac{1}{|\cI^{+}_{i}|} \sum_{i \in \cI^{+}_{i}}^{}{\ell \left(\dd{\bw}{\bx_i} \leq \max \limits_{j \in \cI^{\circ}_i} \dd{\bw}{\bx_j}\right)}\\
&+ &  \frac{1}{|\cI^{\circ}_{i}|} \sum_{i \in \cI^{\circ}_{i}}^{}{\ell \left(\dd{\bw}{\bx_i} \leq \max \limits_{j \in \cI^{-}_i} \dd{\bw}{\bx_j}\right)}
\end{aligned}
\end{equation}
Equipped with the objective  of individual users, we now turn to  the final collaborating ranking  objective as: 
\begin{equation}\label{eqn-primal-social-final}
\begin{aligned}
&\cF(\bW) = \lambda \|\bW\|_{*} \\
& +  \sum_{i = 1}^{n}{\left( \frac{1}{|\cI^{+}_{i}|} \sum_{j \in \cI^{+}_{i}}^{}{\ell\left(\dd{\bw_i}{\bx_{j}} - \|\bX_i^{-}\bw_i\|_{\infty}\right)} \right)}\\
& +   \sum_{i = 1}^{n}{\left(  \frac{1}{|\cI^{+}_{i}|} \sum_{j \in \cI^{+}_{i}}^{}{ \ell\left(\dd{\bw_i}{\bx_{j}} - \|\bX_i^{\circ} \bw_i\|_{\infty}\right)} \right)}\\
&+  \sum_{i=1}^{n}{ \left(\frac{1}{|\cI^{\circ}_{i}|} \sum_{j \in \cI^{\circ}_{i}}^{}{\ell\left(\dd{\bw_i}{\bx_{j}} - \|\bX^{-}_i\bw_i\|_{\infty}\right)}\right)}, 
\end{aligned}
\end{equation}
where  $\bX_i^{\circ} =[\bx_{1}, \ldots, \bx_{n^{\circ}_{i}}]^{\top}$ is the matrix of $n^{\circ}_{i}$ unrated items in  $\cI^{\circ}_{i}$.

\begin{remark} We emphasize that beyond the accuracy considerations of push norm at top of the list,  the  push norm has a clear advantage  to the pairwise ranking models from a computational point of view. In particular,  the push norm has  a linear $O(m)$ dependency on the number of items which is quadratic $O(m^2)$ for pairwise ranking models. 
\end{remark}

\section{The Optimization}\label{sec:optimization}
We now turn to solving the optimization problem in~(\ref{eqn-primal-social-final}). We start by discussing a gradient descent method with shrinkage operator followed by its accelerated version,  and then propose a non-convex  formulation with alternative minimization for more effective optimization of objective in~\alg.

\subsection{Gradient descent with shrinkage operator}

Due to the presence of trace norm of the parameters matrix,  this objective function falls into the general category of composite  optimization, which can be solved by  stochastic gradient or gradient descent methods. In this part we propose a projected gradient decent method to solve the optimization problem. 
First we write the objective as:

\begin{equation}\label{eqn-primal-social-final-1}
\begin{aligned}
\min_{\bW \in \mathbb{R}^{n \times d}}  \cF(\bW) =   \lambda \|\bW\|_{*} + \cL(\bW),
\end{aligned}
\end{equation}
where $\cL(\bW) = \sum_{i=1}^{n}{\cL(\bw_i)}$.

A simple way to solving the above optimization problem is  gradient descent algorithm~\cite{nesterov2013gradient}, which needs to evaluate the gradient of objective at each iteration. To deal with the non-smooth  trace norm $\|\bW\|_{*}$ in the objective, we  first note that the optimization problem in Eq.~(\ref{eqn-primal-social-final-1}) can be reformulated under the framework of proximal regularization or composite gradient mapping~\cite{nesterov2013gradient}. By taking advantage of the composite structure it is possible to retain the same convergence rates of the gradient method for the smooth optimization problems. In particular,  the  optimization problem in~(\ref{eqn-primal-social-final-1}) can be solved iteratively  by:

\begin{equation}\label{eqn-proximal-term}
\begin{aligned}
\bW_t = \arg \min_{\bW} \;&\cL(\bW_{t-1}) + \text{\texttt{tr}} \left((\mtx{W} - \mtx{W}_{t-1})^{\top}\nabla \cL(\mtx{W}_{t-1})\right)  \\
&+\frac{\eta_t}{2}\|\bW - \bW_{t-1}\|_{\rm{F}}^2 +  \lambda \|\bW\|_{*},
\end{aligned}
\end{equation}
where $\{\eta_t\}_{t \geq 1}$ is a sequence of scalar step sizes and $\text{\texttt{tr}}(\cdot)$ is the trace of input matrix. \\

By ignoring the constant terms, Eq.~(\ref{eqn-proximal-term}) can also be rewritten as:
\begin{equation}\label{eqn-proximal-term-8}
\begin{aligned}
\frac{\eta_t}{2} \left\| \bW - \left( \bW_{t-1} - \frac{1}{\eta_{t}} \nabla \cL(\bW_{t-1})\right) \right\|_{\rm{F}}^2 + \lambda \|\bW\|_{*}.
\end{aligned}
\end{equation}

We use the singular value shrinkage operator introduced in~\cite{cai2010singular} to find the optimal solution to Eq.~(\ref{eqn-proximal-term-8}). To this end, consider the singular value decomposition (SVD) of a matrix $\mtx{M} \in \mathbb{R}^{n \times d}$ of rank $r$ as $\mtx{M} = {\bU}  \mtx{\Sigma}   \mtx{V}^*,\quad   \mtx{\Sigma}  = {\diag}(\{\sigma_i\}_{1 \le i \le r}),$
where $\mtx{U}$ and $\mtx{V}$ are respectively $n \times r$ and $d
\times r$ matrices with orthonormal columns, and the singular values
$\sigma_i$ are positive. For a scalar $\tau \in \mathbb{R}_{+}$, define the singular value shrinkage operator  $\mathscr{P}_\tau$ as:
\begin{equation} \label{eqn:DlamM2} 
\mathscr{P}_{\tau}(\mtx{M}):=  \mtx{U}
\mathscr{P}_{\tau}(\mtx{\Sigma}) \mtx{V}^*, \quad
\mathscr{P}_{\tau}(\mtx{\Sigma}) = {\diag}([\sigma_i - \tau]_+\}),
\end{equation}
where $[x]_+$ is the positive part of $x$, namely, $[x]_+ = \max(0, x)$. The shrinkage operator basically  applies a soft-thresholding rule to the singular values of $\mtx{M}$, effectively shrinking these towards zero.

\begin{theorem}[Theorem~2.1,~\cite{cai2010singular}]\label{thm:prox}
For each $\tau \ge 0$ and $\mtx{W} \in \mathbb{R}^{n \times d}$, the  singular value shrinkage operator $\eqref{eqn:DlamM2}$ obeys
\begin{equation}\label{eqn:DlamM}
\mathscr{P}_{\tau}(\mtx{W}) = \arg\min_{\mtx{X}} \left\{\frac12\|\mtx{X}-\mtx{W}\|_{\rm{F}}^2 + \tau\|\mathbf{X}\|_{*}\right\}.
\end{equation}
\end{theorem}

The above theorem shows that the singular value shrinkage operator of a matrix is the solution to Eq.~(\ref{eqn:DlamM}).  Equipped with this result, the  optimization problem in Eq.~(\ref{eqn-proximal-term-8}) can be solved by first computing the SVD of updated $\mtx{W}_{t-1}$ and then applying soft thresholding on the singular values as:
$$\bW_{t} = \mathscr{P}_{\frac{\lambda}{\eta_{t-1}}} \left( \bW_{t-1} - \frac{1}{\eta_{t-1}} \nabla \cL(\bW_{t-1})\right)$$
Now we need to evaluate the gradient of $\cL(\bW)$ at $\bW_{t-1}$. The convex function $\cL(\bW)$ is not differentiable, so we use its subgradient in updating the solutions which can be computed for $i$th parameter vector $\bw_i$,  as follows:
\begin{equation}
\begin{aligned}\label{eqn-grad-u}
 \bg_i &=  {\partial \cL}/{\partial \bw_i} \\
& = \sum_{j \in \cI^{+}_{i}}^{}{\mathbb{I}\left[\dd{\bw_i}{\bx_{j}} - \|\bX_i^{-}\bw_i\|_{\infty} \leq 1\right] \left( \partial \|\bX_i^{-} \bw_i\|_{\infty} - \bx_{j}\right)}\\ 
&+  \sum_{j \in \cI^{+}_{i}}^{}{\mathbb{I}\left[\dd{\bw_i}{\bx_{j}} - \|\bX_i^{\circ}\bw_i\|_{\infty} \leq 1\right] \left( \partial \|\bX_i^{\circ} \bw_i\|_{\infty} - \bx_{j}\right)}\\ 
&+  \sum_{j \in \cI^{\circ}_{i}}^{}{\mathbb{I}\left[\dd{\bw_i}{\bx_{j}} - \|\bX_i^{-}\bw_i\|_{\infty} \leq 1\right] \left( \partial \|\bX_i^{-} \bw_i\|_{\infty} - \bx_{j}\right)}, 
\end{aligned}
\end{equation}
where  $\partial \|\bX_i^{-} \bw_i\|_{\infty}$ is the subdifferential of the function $\|\bX_i^{-} \bw_i\|_{\infty}$ at point $\bw_i$.    Since the subdifferential of the maximum of functions is the convex hull of the union of subdifferentials of the active functions at point $\bw$~\cite{nesterov2004introductory}, we have:
\begin{align*}
\partial \|\bX_i^{-} \bw_i\|_{\infty} &= \partial \max_{1 \leq j \in \cI^{-}_{i}} \dd{\bw_i}{\bx_j} \\
& = \text{conv} \left\{\bx_j| \dd{\bw_i}{\bx_j} = \|\bX_i^{-} \bw_i\|_{\infty}, j \in \cI^{-}_{j} \right\}.
\end{align*}
Then, $\mtx{G} = [\bg_1, \bg_2, \ldots, \bg_n]^{\top} \in \mathbb{R}^{n \times d}$ is a subgradient at  $\bW$, i.e. $\mtx{G} \in \partial \cL(\bW)$.


\begin{remark}
We  note that here, for the ease of exposition, we only consider the non-differentiable convex hinge loss as the surrogate of non-convex due to its computational virtues. However, by using other smooth convex surrogate losses such as smoothed hinge loss or exponential loss, one can apply the accelerated gradient descent methods~\cite{nesterov2004introductory} to solve the optimization problem which results in significantly faster convergence rate compared to the naive gradient descent (i.e, $O(1/\sqrt{\epsilon})$ convergence rate for accelerated method compared to the $O(1/\epsilon^2)$ rate for gradient descent for non-smooth optimization, where $\epsilon$ is the target accuracy). 
\end{remark}

\subsection{Efficient optimization by dropping convexity}
The main computational cost in each iteration of the \alg~algorithm lies in
computing the SVD decomposition of $\bW_k$. An alternative cheaper way to  solving  the optimization problem in Eq.~(\ref{eqn-primal-social-final-1}) is as follows. For a fixed rank of the target parameter matrix $\bW$, say $k$, one can decompose it as $\bW = \bU \bV$. From the equivalence relation between trace norm and the Frobenius of its components in decomposition,
$$ \|\bW\|_*  = \min_{\substack{\bU \in \mathbb{R}^{n \times k}, \bV \in \mathbb{R}^{m \times k}\\ \bW = \bU \bV^{\top} }} \frac{1}{2} \left( \|\bU\|_{\rm{F}}^2 + \|\bV\|_{\rm{F}}^2 \right)$$

we can write the objective in terms of these low rank components as:

\begin{equation}\label{eqn-primal-social-final-2}
\begin{aligned}
\min_{\bU \in \mathbb{R}^{n \times k}, \bV \in \mathbb{R}^{m \times k}}    \frac{\lambda}{2} \left( \|\bU\|_{\rm{F}}^2 + \|\bV\|_{\rm{F}}^2 \right) + \cL(\bU\bV),
\end{aligned}
\end{equation}

These factored optimization problem does not have the explicit trace norm regularization. However, the new formulation is non-convex and potentially subject to stationary points that are not globally optimal. However, despite its non-convexity, the formulation in~Eq.~(\ref{eqn-primal-social-final-2})  is competitive  as compared to trace-norm minimization, while scalability is much better. In particular, the objective is not jointly convex in both $\bU$ and $\bV$ but it is convex in each of them fixing the other one. Therefore, to find a local solution one can  stick to  the standard gradient descent method to find a solution in an iterative manner as follows:
\begin{align*}
& \bU_{t+1} \leftarrow (1 - \lambda \eta_t) \bU_t - \eta_t \nabla_{\bU} \cL\large{|}_{\bU = \bU_t, \bV = \bV_t}, \\
& \bV_{t+1} \leftarrow (1 - \lambda \eta_t) \bV_t - \eta_t \nabla_{\bV} \cL\large{|}_{\bU = \bU_t, \bV = \bV_t}. \\
\end{align*}

\begin{remark} It is remarkable that  the large number of users or items  may cause computational  problems in solving the optimization problem  using GD method. The reason is essentially the fact that computing the gradient at each iteration requires to go through all the users and compute the gradient for pair of items. To alleviate this problem one can utilize stochastic gradient  method~\cite{nemirovski-2009} to solve the optimization problem. The main idea is to choose a fixed subset of pairs for gradient computation instead of all pairs  at \textit{each} iteration or a sample a user at random for gradient computation instead of including all users.  We note that this strategy generates unbiased estimates of the true gradient and makes each iteration of algorithm computationally more efficient compared to the full gradient counterpart. 
\end{remark}


\begin{algorithm}[t]
\center \caption{\texttt{\alg+}}
\begin{algorithmic}[1] \label{alg:taco+}
     \STATE \textbf{input:}  
 $\gamma, \lambda \in \mathbb{R}_{+}$: the regularization parameters,  $\{\eta_t\}_{t \geq 1}$: the sequence of scalar step sizes, and $\mtx{S} \in \mathbb{R}^{n \times n}$: the similarity matrix of usets
	\STATE Initialize $\bW_0 \in \mathbb{R}^{n \times d}$
	\STATE Choose an appropriate step size
	\FOR{$t = 1, \ldots, T$}	
		\STATE Compute the sub-gradient of $\mtx{G}_t \in \partial \cL(\bW_{t})$ using Eq.~(\ref{eqn-grad-u})
		\STATE Compute $\widehat{\mtx{G}}_t = \mtx{G}_t + 2 \gamma \mtx{X}\mtx{X}^{\top} \mtx{W} \mtx{L}$ 
		\STATE $[\mtx{U}_t, \mtx{\Sigma}_t, \mtx{V}_t] \leftarrow \texttt{SVD}(\bW_{t-1} - \frac{1}{\eta_{t-1}}\widehat{\mtx{G}}_t))$
		\STATE $\bW_t \leftarrow \mtx{U}_t \left[ \mtx{\Sigma} - \frac{\lambda}{\eta_{t-1}} \mtx{I}\right]_{+} \mtx{V}_t^{\top}$
	 \ENDFOR
    \STATE  \textbf{output:}  
\end{algorithmic}
\end{algorithm}

\section{Regularization by Exploiting Similarity of Users}\label{sec:taco-users}

In many recommender systems, in addition to handling cold-start items, it would be beneficial to make high quality recommendations to new users.  The new user cold start issue represents a serious problem in recommender systems as it can lead to the loss of new users who decide to stop using the system due to the lack of accuracy in the recommendations received in that first stage in which they have not yet cast a significant number of ratings with which to feed the recommender system's collaborative filtering core.

The above formulation  estimates a parameter vector separately for each user regardless of potential similarities across users. In this section we generalize the proposed ranking model to the setting where the similarity information between users is available.  Let $\bS \in \mathbb{R}^{n \times n}$ be the similarity matrix of users which is inferred from the side information about users such as  social relations between them or explicit users' features.  Let $\bD$ be the diagonal matrix with $D_{ii} = \sum_{j = 1}^{n}{W_{ij}}$, and  $\bL = \bD- \bS$ be the Laplacian matrix. It is natural to require the similar users in the matrix $\bS$ have similarity evaluation  on rating the items~\cite{belkin2006manifold}.  Thus, the new regularization on the parameters of users $\bW$ using the similarity matrix $\bS$ can be achieved by minimizing:
\begin{equation*}\label{}
\begin{aligned}
& \frac{1}{2} \sum_{i = 1}^{m}{ \left( \sum_{j,k = 1}^{n}{S_{jk} \left(\dd{\bw_j}{\bx_i} - \dd{\bw_k}{\bx_i} \right)^2} \right)} \\
& = \frac{1}{2} \sum_{i = 1}^{m} \left( \bx_i^{\top} \bw_j ( \sum_{k}^{}{S_{jk}} ) \bw_j^{\top} \bx_i  - \sum_{j,k}^{}{\bx_i^{\top} \bw_j S_{jk} \bw_{k}^{\top} \bx_i}\right)\\
& = \sum_{i}^{}{ \left( \sum_{j}^{}{\bx_i^{\top} \bw_j S_{jj} \bw_j^{\top} \bx_i}  - \sum_{j,k}^{}{\bx_i^{\top} \bw_j S_{ij} \bw_k^{\top} \bx_i} \right)}\\
& = \sum_{i}^{}{\bx_i^{\top} \left( \bW \bD \bW^{\top} - \bW \bS \bW^{\top}\right)\bx_i}\\
& = \sum_{i}^{}{\bx_i^{\top} \bW \bL \bW^{\top}  \bx_i} =  \text{\texttt{tr}} \left( \bX^{\top} \bW \bL \bW^{\top} \bX\right).
\end{aligned}
\end{equation*}
By plugging the above regularization term in the~Eq.~(\ref{eqn-primal-social-final-1}), we  obtain the following optimization problem:
\begin{equation*}\label{}
\begin{aligned}
\cF(\bW) =  \lambda \|\bW\|_{*} +  \cL(\bW) + \gamma \;\text{\texttt{tr}} \left( \bX^{\top} \bW \bL \bW^{\top} \bX\right), 
\end{aligned}
\end{equation*}

The above optimization problem  can also be solved using the optimization procedure discussed before. The only difference is the gradient computation step which now has an extra term due the introduction  of user's similarity  regularization. Specifically, we have the new gradient $\widehat{\mtx{G}}_t$ computed as: 

$$\widehat{\mtx{G}}_t = \mtx{G}_t + 2 \gamma \mtx{X}\mtx{X}^{\top} \mtx{W} \mtx{L},$$
which replaces the $\mtx{G}_t$ in Algorithm~\ref{alg:taco}. The resulting algorithm, dubbed $\alg+$, is detailed in Algorithm~\ref{alg:taco+}.

\section{Experiments}\label{sec:results}

In this section, we conduct exhaustive experiments to demonstrate the merits and advantages of the proposed algorithm. We conduct our experiments on three well-known datasets MovieLens, Amazon and CiteULike. We investigate how the proposed \alg~performs in comparison to the state-of-the-art methods. In the following, first we introduce the datasets that we use in our experiments and then the metrics that we employ to evaluate the results, followed by our detailed experimental results on the real datasets.

In the following subsections, we intend to answer these key questions:
\begin{itemize}
\item \textbf{Ranking versus rating:} How  does  learning optimization for a ranking based loss function  affect  the  performance of recommending versus the square loss function?

\item \textbf{Employing missing ratings:} How does employing the missing ratings could help in making more accurate recommendations?

\item \textbf{Dealing with cold-start items:} How  does  the  proposed  algorithm, with incorporating side information of items, perform in comparison to the  state-of-the-art algorithms to deal with cold-start items?

\end{itemize}

\subsection{Datasets}~We use the following well known datasets to evaluate the performance of \alg{}:\\ 

\begin{itemize}
\item \textbf{\dsML.} We used \dsML{} which is a dataset extracted from the IMDB and the MovieLens $1$M  datasets by mapping the MovieLens and IMDB and collecting the movies that have plots and keywords. The rating values are 10 discrete numbers ranging from $1$ to $10$ and the rating were made binary by treating all the ratings greater than $5$ as $+1$ and below $5$ as $-1$.  \\ 
\item \textbf{Amazon.} We used the dataset of best-selling books and their ratings in Amazon. Each book has a one or two paragraphs of textual description, which has been used to have a set of features of the books. Ratings can be integers numbers from $1$ to $5$. The ratings were also made binary by treating all the ratings greater or equal to $3$ as $+1$ and below $3$ as $-1$.  \\
\item \textbf{CiteULike.} It is an online free service for managing and discovering scholarly references. Users can add those articles that they are interested in to their libraries. Collected articles in a user's library will be considered as relevant items for that user. This dataset does not have explicit irrelevant items and was chosen to illustrate the effect of considering missing data while only having relevant itmes.\\
\end{itemize}

For all above datasets, the description about the items were tokenized and after removing the stop words, the rest of the words were stemmed. Then those words that have been appeared in less than $20$ items and more that $20\%$ of the items were also removed ~\cite{sharmafeature2015b}. At the end, the TF-IDF was applied on the remaining words and the TF-IDF scores represented the features of the items. The statistics of the datasets are given in Table \ref{statisticsOfRealDatasets}. As it is shown in Table \ref{statisticsOfRealDatasets}, all these three datasets have high dimensional feature space.

\begin{table*}
\centering
\caption{Statistics of real datasets used in our experiments.}
\label{statisticsOfRealDatasets}
\begin{tabular} { l || c || c || c }
Statistics & \dsML{} & \dsAM{} & \dsCU{} \\
\hline \hline
Number of users    & 2,113     & 13,097     &   3,272  \\
\hline
Number of items    & 8,645    & 11,077  & 21,508\\
\hline
Number of ratings  & 739,973  & 175,612  &  180,622\\
\hline
Number of features   & 8,744  &  5,766  &  6,359\\
\hline
Average number of ratings by users   &  350.2 &  13.4 &  55.2\\
\hline
Average number of ratings for items  & 85.6  & 13.4  & 55.2\\
\hline
Density  & 4.05\%  & 0.12\%  & 0.13\% \\
\hline
\end{tabular}
\end{table*}


\subsection{Metrics}~
We adopt the widely used metrics, Discounted Cumulative Gain at $n$ and Recall at $n$, for assessing the performance of our and baseline algorithms. For each user $u$, given an item  $i$, let $s_k$ be the relevance score of the item ranked at position $k$, where $s_k = 1$ if the item is relevant to the user $u$ and $s_k = 0$ otherwise. Now, given the list of top-$n$ item recommendations for user $u$, Discounted Cumulative Gain at $n$, is defined as: 
$$
{\rm{DCG_u}}@n = s_1 + \sum_{k=2}^{n}{\frac{s_k}{\log_2 (k)}}
$$
If we divide the ${\rm{DCG_u}}@n$ by its maximum value, we get the ${\rm{NDCG_u}}@n$ value. 
Given the list of top-$n$ item recommendations for each user $u$, Recall at $n$ will count the number of relevant items appeared in the recommendation list divided by the size of the list. Recall at $n$ is defined as:
$$
{\rm{REC_u}}@n =  \frac{| \{\text{relevant items to $u$}\} \cap \{\text{top-$n$ recommended items}\} |}{|\{\text{top-$n$ recommended items}\}|}  
$$
DCG$@n$, ${\rm{NDCG_u}}@n$ and REC$@n$ will be computed for each user and then will be averaged over all users.


\subsection{Methodology}~
Given the partially observed rating matrix, we transformed the observed ratings of all datasets from a multi-level relevance scale to a two-level scale $(+1, -1)$ while $0$ is considered for unobserved ratings.
We randomly selected $60\%$ of the observed ratings for training and $20\%$ for validation set and consider the remaining $20\%$ of the ratings as our test set. To better evaluate the results, we performed a 3-fold-cross validation and reported the average value for our results.



\subsection{Baseline Algorithms}~The proposed~\alg~algorithm is compared to the following algorithms:

\begin{itemize}
\item \textbf{Matrix Factorization (MF) \cite{srebro2004maximum}:} Is a matrix completion method that factorizes the incomplete observed matrix and completes the matrix using the unveiled  latent features.

\item \textbf{Matrix Factorization with Side Information (KPMF)~\cite{zhou2012kernelized}:} 
Is a matrix completion based algorithm, which incorporates external side information of the users or items into the matrix factorization process. It imposes a Gaussian Process prior over all rows of the matrix, and the learned model explicitly captures the underlying correlation among the rows.

\item \textbf{Decoupled Completion and Transduction (DCT)~\cite{barjasteh2015cold}:} Is a matrix factorization based algorithm that decouples the completion and transduction stages and exploits the similarity information among users and items to complete the (rating) matrix.

%
%
%
%
%
%

\item \textbf{Feature Based Factorized Bilinear Similarity Model (FBS) ~\cite{sharmafeature2015b}:} This algorithm uses bilinear model to capture pairwise dependencies between the features. In particular, this model accounts for the interactions between the different item features. 

\item \textbf{Collaborative User-specific Feature-based Similarity Models (CUFSM):} By using the history of ratings for users, it learns personalized user model across the dataset. This method is one of the best performing collaborative latent factor based model~\cite{elbadrawy2015user}.
 \item \textbf{Regression based Latent Factor Model (RLF):\footnote{The implementation of this method is available in LibFM library~\cite{rendle2012factorization}.}} This method incorporates the features of items in factorization process by transforming the features to the latent space using linear regression~\cite{agarwal2009regression}. If the learning method is Markov Chain Monte Carlo, we name it RLF-MCMC. 
 \item \textbf{Cosine Similarity Based Recommender (CSR):}  Using the similarity between features of items, the preference score of a user on an item will be estimated.
 \end{itemize}

We would like to mention that as baseline  algorithms we only consider state-of-the art methods that are able to exploit the side information about items.

\subsection{\alg{} vs. rating}
Many different algorithms are trying to provide recommendations to users such that the predicted rating values be very close to the actual rates that users would provide. These algorithms try to minimize the error between the predicted values and actual rating values by minimizing Mean Squared Error (MAE) or Root Mean Square Error (RMSE) or etc.  \\
Then, due to the fact that users tend to only care about the top of their recommendation list, predicting a ranking of items of interest instead of ratings became the main focus of recent works \cite{christakopoulou2015collaborative}. In this section we compare the results of \alg{} with those state-of-the-art algorithms that try to predict ratings for unrated items. Among the state-of-the-art algorithms, we chose a diverse set of algorithms, which are Matrix Factorization, Matrix Factorization with Side Information, Decoupled Completion and Transduction and Regression Based Latent Factor Model. Table \ref{ResultsOfRankingVsRating} shows the NDCG value of top 10 items of recommendation list. It shows that \alg{} outperformed all other rating prediction based algorithms in terms of NDCG measure. The results confirm the effectiveness of providing the ranks of items rather than their ratings.

\begin{table*}
\centering
\caption{Results of comparing rankings methods versus ratings methods on  \dsML{}. $\lambda$ is regularization parameter, $h$ is dimension of latent features, $T$ is the number of iterations, $\eta $ is learning rate and $\sigma$ is the standard deviation.}
\label{ResultsOfRankingVsRating}
\begin{tabular}{ l | c | c  }
\multirow{2}{*}{\textbf{Algorithms}} &\multirow{2}{*}{\textbf{Hyperparameters}}& \multirow{2}{*}{\textbf{ NDCG$@10$}} \\ 
 &  & \\
           \hline\hline
           
  MF     &$\eta=0.003, h=10,  \sigma = 0.4$&  0.0947 \\ \cline{1-3}

 KPMF &$\eta=0.003, h=10,  \sigma = 0.4$ & 0.1005 \\ \cline{1-3}

 DCT &$      h  = 10$& 0.0095 \\ \cline{1-3}

 RLF-MCMC & $ T=100, \sigma= 0.1 $ & 0.0248    \\ \cline{1-3}

 \alg{} &$\lambda = 0.55, h =10, T=100$ & \textbf{0.265}    \\ \cline{1-3} \hline \hline 

\end{tabular}
\end{table*}


\subsection{Robustness to not missing  at random ratings}
In this section we compare the effect of incorporating the unobserved ratings in our learning in comparison with excluding them from our learning. Most of the methods in the literature ignore the unobserved ratings and train their model only base on observed ratings. By incorporating the unrated items in ranking, our method can limit the bias caused by learning solely based on the observed ratings and consequently deals with the not missing at random issue of ratings. Table \ref{ResultsOfNotMissingAtRandom} shows results of comparing these two scenarios for \alg{} on \dsML{}. In order to see the difference between these two scenarios, we considered 70\% of the ratings for training and 30\% for test to have more ground truth for our testing. Table \ref{ResultsOfNotMissingAtRandom} shows the NDCG@5, 10,15 and 20 for both scenarios and it shows that incorporating the unobserved ratings causes to improve the accuracy of recommendation list. Hence, the NDCG values for top 5, 10, 15 and 20 items improved when unrated items were included as part of the training process.

\begin{table*}
\centering
\caption{Results of employing missing ratings versus ignoring them on  \dsML{}. $\lambda=0.6$ is regularization parameter, $h=10$ is dimension of latent features, $T=100$ is the number of iterations. }
\label{ResultsOfNotMissingAtRandom}
\begin{tabular}{ c | c | c  | c | c  }
\multirow{2}{*}{\textbf{Algorithm: \alg{}}} & \multirow{2}{*}{\textbf{ NDCG@5}}& \multirow{2}{*}{\textbf{ NDCG@10}}& \multirow{2}{*}{\textbf{ NDCG@15}}& \multirow{2}{*}{\textbf{ NDCG@20}} \\ 
 &  & && \\
           \hline\hline
           

Observed ratings     &1.1690   &   2.2218   & 2.8362  &  3.2849      \\ \cline{1-5} 
Observed + missing ratings  &\textbf{1.1794}   &   \textbf{2.2405}   & \textbf{2.8585}  &  \textbf{3.3096}  \\ \hline \hline 

\end{tabular}
\end{table*}

\subsection{Dealing with cold-start items}

We now turn to  evaluating the effectiveness of \alg{} for cold-start recommendation.
To do so, we randomly selected $60\%$ of the items as our training items and $20\%$ for validation set and considered the remaining $20\%$ of the items as our test set.
In this scenario, baseline algorithms that are used for comparison are CSR,  FBS, CUFSM and RLF. For the experiments, we used \dsML{}, \dsAM{} and \dsCU{} datasets. Table \ref{ResultsOfColdStartScenario} shows the measurement results of applying mentioned algorithms on these datasets. For each test, the parameters' values producing the best ranking on the validation set  were selected to be used and reported. As it can be seen from the results in Table~\ref{ResultsOfColdStartScenario}, the proposed ~\alg~ algorithm outperformed all other baseline algorithms and provided a recommendations with higher quality in comparison to other methods. We can also see from the results of Table~\ref{ResultsOfColdStartScenario} that for the ML-IMDB dataset, the improvement in terms of REC@10 is significant compared to other datasets. Since the density of this dataset is much higher than other two datasets, this observation indicates that our method is more effective in utilizing side information compared to other methods.
These results demonstrate the effectiveness of \alg{} in comparison with other state-of-the-art algorithms. \alg{} was able to outperform other state-of-the-art algorithms by considering the missing data and focusing on top of the recommendation list for cold-start items.

%
%

%


%

%
%
%
%
%
%

\begin{table*}
\centering
\caption{Results on cold-start items. $\lambda$, $\mu_1$ and $\beta$ are regularization parameters, $h$ is dimension of latent features, $l$ is the number of similarity functions and $T$ is the number of iterations.}
\label{ResultsOfColdStartScenario}
\begin{tabular}{c | c | c | c | c   }
&\multirow{2}{*}{\textbf{Algorithms}} &\multirow{2}{*}{\textbf{Hyperparameters}}& \multirow{2}{*}{\textbf{ DCG$@10$}}&\multirow{2}{*}{\textbf{REC$@10$}}  \\ 
  &&  & &  \\
           \hline\hline
           
\multirow{5}{*}{\rotatebox{90}{\textbf{\dsML{}}}}&  CSR  &---&  0.1282 & 0.0525 \\ \cline{2-5}

& RLF &$h = 15$ & 0.0455 & 0.0155 \\ \cline{2-5}

& CUFSM &$l=1, \mu_1=0.005$& 0.2160 & 0.0937\\ \cline{2-5}

& FBS &$\lambda=0.01, \beta=0.1, h=5$ & 0.2270 & 0.0964  \\ \cline{2-5}

& \alg{} &$\lambda = 0.6, h =10, T=200$ & \textbf{0.2731}  & \textbf{0.2127}  \\ \cline{1-5} \hline \hline 

\multirow{5}{*}{\rotatebox{90}{\textbf{\dsAM{}}}}&  CSR  &---&  0.0228 & 0.1205\\ \cline{2-5}

& RLF &$h = 30$ & 0.0076 & 0.0394 \\ \cline{2-5}

& CUFSM &$l=1, \mu_1=0.25$& 0.0282 & 0.1376  \\ \cline{2-5}

& FBS &$\lambda=0.1, \beta=1, h=1$ & 0.0284 & 0.1392 \\ \cline{2-5}

& \alg{} &$\lambda = 0.6, h =10, T=200$ & \textbf{0.1195} & \textbf{0.1683} \\ \cline{1-5} \hline \hline 

\multirow{5}{*}{\rotatebox{90}{\textbf{\dsCU{}}}}&  CSR  &---& 0.0684 & 0.1791  \\ \cline{2-5}

& RLF &$h = 75$ & 0.0424 & 0.0874  \\ \cline{2-5}

& CUFSM &$l=1, \mu_1=0.25$& 0.0791 & 0.2017\\ \cline{2-5}

& FBS &$\lambda=0.25, \beta=10, h=5$ & 0.0792 & 0.2026  \\ \cline{2-5}

& \alg{} &$\lambda = 0.6, h =10, T=200$ &  \textbf{0.0920} &  \textbf{0.2243}  \\ \cline{1-5} \hline \hline 
\end{tabular}
\end{table*}

\section{Conclusions}\label{sec:conclusions}

In this paper we introduced a   {semi-supervised collaborative ranking} model  by leveraging side information about {both observed and missing ratings} in collaboratively learning the ranking model.  In the learned model, unrated items are conservatively pushed after the relevant  and before the relevant items in the ranked list of items for each individual user. This crucial difference greatly boosts the performance and limits the bias caused by learning only from sparse non-random observed ratings. The proposed algorithm is compared with seven baseline algorithms on three real world datasets that demonstrated the effectiveness of  proposed algorithm in addressing cold-start problem and mitigating the data sparsity problem, while being robust to sampling of missing ratings.

This work leaves few interesting directions as future work. First, we would like to investigate the performance of the proposed \alg~algorithm when side information about users is also available using the graph regularization idea discussed in Section~\ref{sec:taco-users}. Second, we would like to empirically evaluate the performance of the optimization method derived by dropping the convexity in future. Also, we have largely ignored the case of differentiable smooth surrogate convex loss functions in this work and it would be interesting to consider smooth alternatives and apply the accelerated optimization methods for faster convergence. Moreover,  the scalability analysis of of proposed algorithm on large datasets using stochastic optimization methods is also worthy of investigation. Finally, we believe  there are   still many open questions relating to non-random nature of rating information in many real applications.
 
\bibliographystyle{abbrv}
\bibliography{colloborative-ranking@arXiv}

\begin{thebibliography}{10}

\bibitem{100}
G.~Adomavicius and A.~Tuzhilin.
\newblock Toward the next generation of recommender systems: A survey of the
  state-of-the-art and possible extensions.
\newblock {\em IEEE Transactions on Knowledge and Data Engineering},
  17(6):734--749, 2005.

\bibitem{agarwal2009regression}
D.~Agarwal and B.-C. Chen.
\newblock Regression-based latent factor models.
\newblock In {\em SIGKDD}, pages 19--28. ACM, 2009.

\bibitem{agarwal2011infinite}
S.~Agarwal.
\newblock The infinite push: A new support vector ranking algorithm that
  directly optimizes accuracy at the absolute top of the list.
\newblock In {\em SDM}, pages 839--850. SIAM, 2011.

\bibitem{balakrishnan2012collaborative}
S.~Balakrishnan and S.~Chopra.
\newblock Collaborative ranking.
\newblock In {\em ACM WSDM}, pages 143--152. ACM, 2012.

\bibitem{barjasteh2015cold}
I.~Barjasteh, R.~Forsati, F.~Masrour, A.-H. Esfahanian, and H.~Radha.
\newblock Cold-start item and user recommendation with decoupled completion and
  transduction.
\newblock In {\em Proceedings of the 9th ACM Conference on Recommender
  Systems}, pages 91--98. ACM, 2015.

\bibitem{belkin2006manifold}
M.~Belkin, P.~Niyogi, and V.~Sindhwani.
\newblock Manifold regularization: A geometric framework for learning from
  labeled and unlabeled examples.
\newblock {\em The Journal of Machine Learning Research}, 7:2399--2434, 2006.

\bibitem{burges1998tutorial}
C.~J. Burges.
\newblock A tutorial on support vector machines for pattern recognition.
\newblock {\em Data mining and knowledge discovery}, 2(2):121--167, 1998.

\bibitem{cai2010singular}
J.-F. Cai, E.~J. Cand{\`e}s, and Z.~Shen.
\newblock A singular value thresholding algorithm for matrix completion.
\newblock {\em SIAM Journal on Optimization}, 20(4):1956--1982, 2010.

\bibitem{candes2010power}
E.~J. Cand{\`e}s and T.~Tao.
\newblock The power of convex relaxation: Near-optimal matrix completion.
\newblock {\em Information Theory, IEEE Transactions on}, 56(5):2053--2080,
  2010.

\bibitem{christakopoulou2015collaborative}
K.~Christakopoulou and A.~Banerjee.
\newblock Collaborative ranking with a push at the top.
\newblock In {\em WWW}, pages 205--215. International World Wide Web
  Conferences Steering Committee, 2015.

\bibitem{cremonesi2010performance}
P.~Cremonesi, Y.~Koren, and R.~Turrin.
\newblock Performance of recommender algorithms on top-n recommendation tasks.
\newblock In {\em ACM RecSys}, pages 39--46. ACM, 2010.

\bibitem{devooght2015dynamic}
R.~Devooght, N.~Kourtellis, and A.~Mantrach.
\newblock Dynamic matrix factorization with priors on unknown values.
\newblock In {\em ACM SIGKDD}, pages 189--198. ACM, 2015.

\bibitem{elbadrawy2015user}
A.~Elbadrawy and G.~Karypis.
\newblock User-specific feature-based similarity models for top-n
  recommendation of new items.
\newblock {\em ACM Transactions on Intelligent Systems and Technology (TIST)},
  6(3):33, 2015.

\bibitem{jarvelin2000ir}
K.~J{\"a}rvelin and J.~Kek{\"a}l{\"a}inen.
\newblock Ir evaluation methods for retrieving highly relevant documents.
\newblock In {\em ACM SIGIR}, pages 41--48. ACM, 2000.

\bibitem{koren2009matrix}
Y.~Koren, R.~Bell, and C.~Volinsky.
\newblock Matrix factorization techniques for recommender systems.
\newblock {\em Computer}, (8):30--37, 2009.

\bibitem{lee2014local}
J.~Lee, S.~Bengio, S.~Kim, G.~Lebanon, and Y.~Singer.
\newblock Local collaborative ranking.
\newblock In {\em Proceedings of the 23rd international conference on World
  wide web}, pages 85--96. ACM, 2014.

\bibitem{li2014top}
N.~Li, R.~Jin, and Z.-H. Zhou.
\newblock Top rank optimization in linear time.
\newblock In {\em Advances in Neural Information Processing Systems}, pages
  1502--1510, 2014.

\bibitem{liu2009learning}
T.-Y. Liu.
\newblock Learning to rank for information retrieval.
\newblock {\em Foundations and Trends in Information Retrieval}, 3(3):225--331,
  2009.

\bibitem{marlin2009collaborative}
B.~M. Marlin and R.~S. Zemel.
\newblock Collaborative prediction and ranking with non-random missing data.
\newblock In {\em RecSys}, pages 5--12. ACM, 2009.

\bibitem{missing-at-random2007}
B.~M. Marlin, R.~S. Zemel, S.~T. Roweis, and M.~Slaney.
\newblock Collaborative filtering and the missing at random assumption.
\newblock In {\em UAI}, pages 267--275, 2007.

\bibitem{nemirovski-2009}
A.~Nemirovski, A.~Juditsky, G.~Lan, and A.~Shapiro.
\newblock Robust stochastic approximation approach to stochastic programming.
\newblock {\em SIAM Journal on Optimization}, 19(4):1574--1609, 2009.

\bibitem{nesterov2004introductory}
Y.~Nesterov.
\newblock {\em Introductory lectures on convex optimization}, volume~87.
\newblock Springer Science \& Business Media, 2004.

\bibitem{nesterov2013gradient}
Y.~Nesterov.
\newblock Gradient methods for minimizing composite functions.
\newblock {\em Mathematical Programming}, 140(1):125--161, 2013.

\bibitem{park2009pairwise}
S.-T. Park and W.~Chu.
\newblock Pairwise preference regression for cold-start recommendation.
\newblock In {\em RecSys}, pages 21--28. ACM, 2009.

\bibitem{rendle2012factorization}
S.~Rendle.
\newblock Factorization machines with libfm.
\newblock {\em ACM Transactions on Intelligent Systems and Technology (TIST)},
  3(3):57, 2012.

\bibitem{rudin2009p}
C.~Rudin.
\newblock The p-norm push: A simple convex ranking algorithm that concentrates
  at the top of the list.
\newblock {\em The Journal of Machine Learning Research}, 10:2233--2271, 2009.

\bibitem{saveski2014item}
M.~Saveski and A.~Mantrach.
\newblock Item cold-start recommendations: learning local collective
  embeddings.
\newblock In {\em RecSys}, pages 89--96. ACM, 2014.

\bibitem{schein2002methods}
A.~I. Schein, A.~Popescul, L.~H. Ungar, and D.~M. Pennock.
\newblock Methods and metrics for cold-start recommendations.
\newblock In {\em SIGIR}, pages 253--260. ACM, 2002.

\bibitem{sharmafeature2015b}
M.~Sharma, J.~Zhou, J.~Hu, and G.~Karypis.
\newblock Feature-based factorized bilinear similarity model for cold-start
  top-n item recommendation.
\newblock In {\em SDM}, 2015.

\bibitem{shi2010list}
Y.~Shi, M.~Larson, and A.~Hanjalic.
\newblock List-wise learning to rank with matrix factorization for
  collaborative filtering.
\newblock In {\em ACM RecSys}, pages 269--272. ACM, 2010.

\bibitem{shi2014collaborative}
Y.~Shi, M.~Larson, and A.~Hanjalic.
\newblock Collaborative filtering beyond the user-item matrix: A survey of the
  state of the art and future challenges.
\newblock {\em ACM Computing Surveys (CSUR)}, 47(1):3, 2014.

\bibitem{sindhwani2010one}
V.~Sindhwani, S.~S. Bucak, J.~Hu, and A.~Mojsilovic.
\newblock One-class matrix completion with low-density factorizations.
\newblock In {\em ICDM}, pages 1055--1060. IEEE, 2010.

\bibitem{srebro2004maximum}
N.~Srebro, J.~Rennie, and T.~S. Jaakkola.
\newblock Maximum-margin matrix factorization.
\newblock In {\em Advances in neural information processing systems}, pages
  1329--1336, 2004.

\bibitem{steck2010training}
H.~Steck.
\newblock Training and testing of recommender systems on data missing not at
  random.
\newblock In {\em KDD}, pages 713--722. ACM, 2010.

\bibitem{steck2015gaussian}
H.~Steck.
\newblock Gaussian ranking by matrix factorization.
\newblock In {\em Proceedings of the 9th ACM Conference on Recommender
  Systems}, pages 115--122. ACM, 2015.

\bibitem{volkovs2012collaborative}
M.~Volkovs and R.~S. Zemel.
\newblock Collaborative ranking with 17 parameters.
\newblock In {\em Advances in Neural Information Processing Systems}, pages
  2294--2302, 2012.

\bibitem{weimer2007maximum}
M.~Weimer, A.~Karatzoglou, Q.~V. Le, and A.~Smola.
\newblock Maximum margin matrix factorization for collaborative ranking.
\newblock {\em NIPS}, 2007.

\bibitem{zhou2012kernelized}
T.~Zhou, H.~Shan, A.~Banerjee, and G.~Sapiro.
\newblock Kernelized probabilistic matrix factorization: Exploiting graphs and
  side information.
\newblock In {\em SDM}, volume~12, pages 403--414. SIAM, 2012.

\end{thebibliography}
\end{document}